\documentclass[letterpaper, 10 pt, conference]{ieeeconf}  

\IEEEoverridecommandlockouts                              

\usepackage{graphics} 
\usepackage{epsfig} 
\usepackage{amsmath} 

\usepackage{graphicx}
\usepackage{subcaption}
\usepackage{multirow}

\title{\LARGE \bf
Resolving Positional Ambiguity in Dialogues by Vision-Language Models for Robot Navigation
}

\author{Kuan-Lin Chen, Tzu-Ti Wei, Li-Tzu Yeh, Elaine Kao$^*$, Yu-Chee Tseng, and Jen-Jee Chen
\thanks{The authors are with the College of Artificial Intelligence, National Yang Ming Chiao Tung University (NYCU), TAIWAN.
$^*$Elaine Kao worked as an intern at NYCU.}
}

\begin{document}
\maketitle
\thispagestyle{empty}
\pagestyle{empty}

\begin{abstract}
We consider an autonomous navigation robot that can accept human commands through natural language to provide services in an indoor environment. These natural language commands may include time, position, object, and action components. However, we observe that the positional components within such commands usually refer to objects in the environment that may contain different levels of positional ambiguity. For example, the command ``\texttt{Go to the chair!}'' may be ambiguous when there are multiple chairs of the same type in a room. In order to disambiguate these commands, we employ a large language model and a large vision-language model to conduct multiple turns of conversations with the user. We propose a two-level approach that utilizes a vision-language model to map the meanings in natural language to a unique object ID in images and then performs another mapping from the unique object ID to a 3D depth map, thereby allowing the robot to navigate from its current position to the target position. To the best of our knowledge, this is the first work linking foundation models to the positional ambiguity issue.
\end{abstract}


\section{Introduction}

With the advancements in sensor and computer vision, significant progress in robotics has been made in autonomous mapping \cite{leonard1991simultaneous,stachniss2009robotic} and obstacle avoidance \cite{borenstein1989real,borenstein1991histogramic,pandey2017mobile}. Machine learning techniques for LiDAR and cameras have been developed for better decision-making in dynamic environments \cite{soori2023artificial}, which have led to more reliable and efficient navigation techniques for robots in various fields such as delivery, manufacturing, and exploration. These advancements include simultaneous localization and mapping (SLAM) \cite{alsadik2021simultaneous, gobhinath2021simultaneous}, enhanced sensor fusion \cite{alatise2020review}, and real-time object recognition \cite{maturana2015voxnet,guo2021real}, which largely improve robots' navigation capabilities in complex environments.

On the other hand, recent advances in artificial intelligence have led to significant breakthroughs in large language models (LLM) and large multi-modal models (LMM), such as GPT~\cite{achiam2023gpt} and BERT \cite{devlin2018bert}, which revolutionize natural language understanding and generation in various domains. These advances include generating coherent and contextually relevant text, understanding complex instructions, and even reasoning on language-based tasks. In particular, LMMs infer and align information from vision, language, and even sensory data \cite{fritsch2003multi, tang2023comparative, huang2023visual}, significantly moving forward human-to-robot and robot-to-physical world interactions. 

Despite advancements in autonomous robot navigation, there remains a large gap between humans' dialogues with robots and robots' perception of a physical space. More specifically, the cutting-edge technologies already allow a robot to precisely scan a space with deep comprehension and construct a 3D map. However, when a human interacts with a robot by natural language, the dialogues may easily contain ambiguity that confuses a robot. For example, the commands
\begin{quote}
\texttt{Take the cup on the table.}
\\
\texttt{Go to the chair nearby the window.}
\end{quote}
would confuse a robot when there are two cups/chairs meeting the criteria. Therefore, second commands like
\begin{quote}
\texttt{The cup next to the vase.}
\\
\texttt{The chair facing the window.}
\end{quote}
may clarify the ambiguity. Then the robot can map the unique object on its LiDAR map, navigate to the right position, and then perform the task. 

This work aims to bridge the gap between the perception of a robot on an environment and the natural language commands given by humans. We consider natural language commands that consist of four components: time, position, target object, and required action. We focus on the ambiguity issue that may exist in the positional component. In the above examples, the robot needs to infer the right cup/chair in order to infer a target position. Whenever there is an ambiguity, a second or a third dialogue can be initiated to clarify the ambiguity. We leverage the multi-dialogue capability of the state-of-the-art LLM and vision-language model (VLM) to resolve such ambiguity, thus enabling a robot to precisely map to a target position in its internal 3D physical map. 

\begin{figure}
    \centering
    \includegraphics[width=0.35\textwidth]{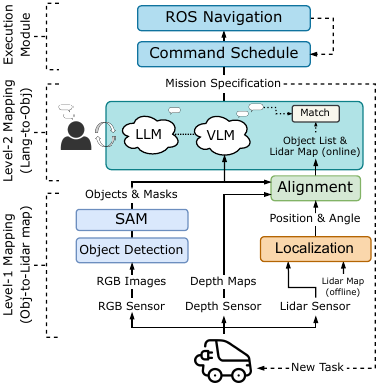}
    \caption{The robot navigation framework.}
    \label{fig:architecture}
\end{figure}

Fig. \ref{fig:architecture} outlines our robot navigation framework. First, the \emph{Level-1 Mapping} conducts 3D scanning and object detection on the space to obtain an accurate LiDAR map with all objects uniquely mapped to the LiDAR map. Also, a unique ID is assigned to each object. Second, the \emph{Level-2 Mapping} leverages the power of LLM and VLM, which takes user's dialogues and photo snapshots from the robot's current position as inputs, to map the positional meanings in the dialogues to a unique object in photos. Especially on these photos, annotations of object IDs are given in bounding box forms for VLM to clarify positional ambiguity. Through the above two-level mapping, the \emph{Execution Module} then instructs the robot to navigate to the target position according to the 3D LiDAR map.

Sec. 2 reviews some related work. Sec. 3 presents our detailed method. Experiment results are shown in Sec. 4. Sec. 5 concludes this paper.

\section{Related Work}
\subsection{Vision-and-Language Navigation (VLN)}

In recent years, Vision-and-Language Navigation (VLN) has garnered increasing attention. Previous approaches mainly focused on data augmentation \cite{liu2021vision,li2022envedit}, memory mechanisms \cite{chen2021history,wang2021structured}, and pre-training \cite{hao2020towards,guhur2021airbert}. These methods emphasized how to effectively utilize simulator data during training to enhance simulation performance, but overlooked their applications in the real world. Recently, the development of LLMs has advanced research on zero-shot vision-and-language navigation agents, enabling these agents to navigate in real-world environments. LM-Nav \cite{shah2023lm} utilizes GPT-3 to navigate through real-world topological maps; however, this approach struggles to represent spatial relationships between objects, potentially leading to the loss of detailed information. In contrast, CoW \cite{gadre2023cows} and VLMaps \cite{huang2023visual} use top-down semantic maps to model the navigation environment, which allow for a more accurate representation of spatial relationships. However, these methods are limited by predefined semantic labels, restricting their semantic concepts. Unlike these approaches, NavGPT \cite{zhou2024navgpt} constructs a vision-language navigation framework that translates visual scene semantics into prompts, allowing an LLM to directly execute VLN tasks. However, since NavGPT relies solely on text descriptions as input, its performance is constrained by the quality of the language descriptions of visual scenes. 

Our approach leverages the power of VLM to jointly process visual inputs and language cues and employs multi-turn dialogues to resolve positional ambiguities in human commands. Our level-2 mapping takes advantage of VLM's vision-language alignment capability to identify a unique object, thereby improving the accuracy of VLN.

\begin{figure}
    \centering
    \includegraphics[width=0.35\textwidth]{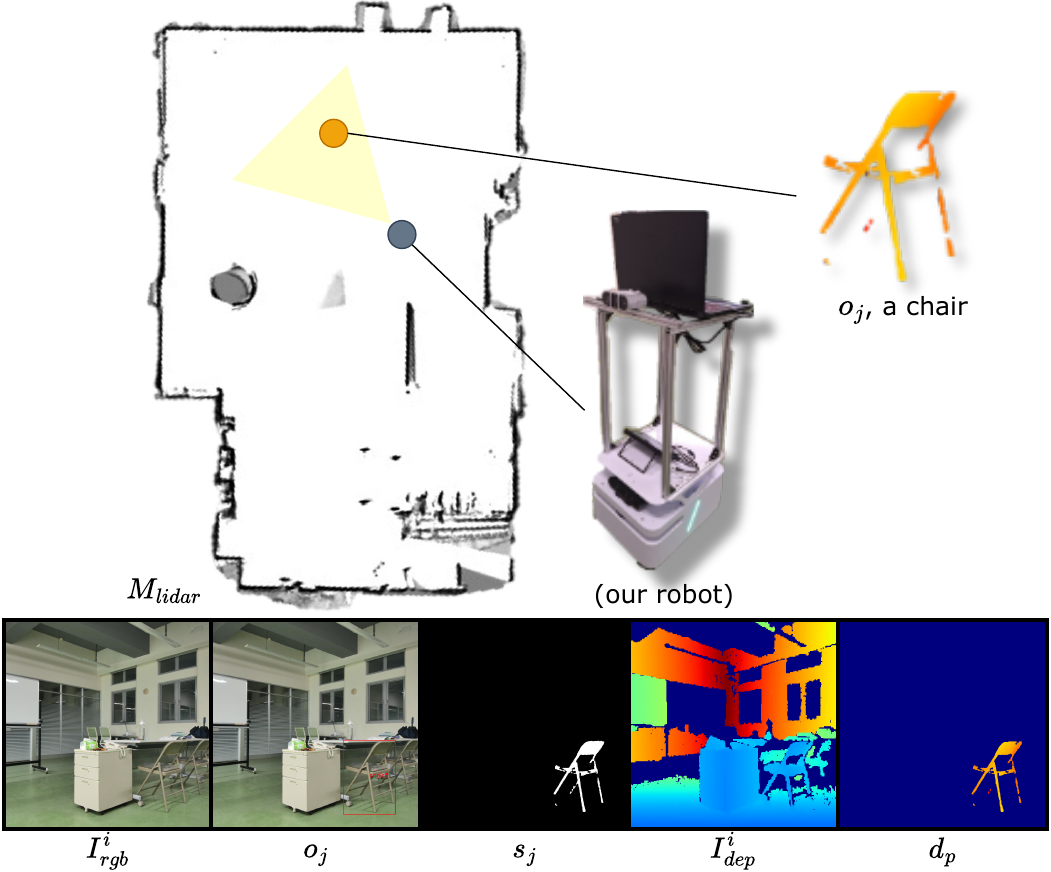}
    \caption{An example of level-1 mapping.}
    \label{fig:level1_example}
\end{figure}

\subsection{LLM and LMM}

The breakthroughs in LLM have demonstrated broad and impressive capabilities across various domains, such as summarization, code generation, and task planning. Inspired by the fine-tuning of LLMs with instructions, researchers have made progress in integrating visual instruction fine-tuning, thereby enhancing VLMs' ability to follow instructions. Vision-language models like GPT-4V \cite{achiam2023gpt}, LLaVA \cite{liu2024visual}, and InstructBLIP \cite{dai2023instructblipgeneralpurposevisionlanguagemodels} have excelled in responding to image content, detailed semantics, and image-to-text artistic creation. These advancements inspired us to leverage VLM for precise navigation.

\section{Position Disambiguation Methodology}

Fig. \ref{fig:architecture} shows our navigation framework. The robot is equipped with a LiDAR and a RGB-D camera. However, the user only uses natural language to communicate with the robot via a speech-to-text interface. We assume that before the robot starts to navigate, it has already explored and scanned the space to construct an offline 2D LiDAR map $M_{lidar}$. To assign a mission to the robot, the user needs to exchange one or multiple dialogues with our system. Missions can be assigned one-by-one and put into a scheduler. Below, we discuss the initiation of one mission.

\begin{figure*}[t]
    \centering
    \includegraphics[width=0.85\textwidth]{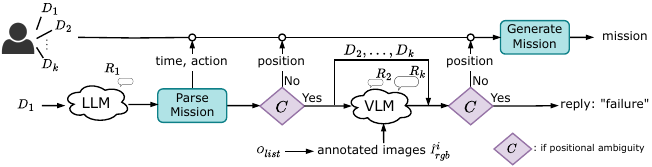}
    \caption{The language-to-object (level-2) mapping module.}
    \label{fig:dialogue}
\end{figure*}

\subsection{Level-1 Mapping: Object-to-LiDAR Map}

During navigation, the robot will continuously execute two tasks: (i) position itself in the LiDAR map and (ii) collect object information in its current environment. Task (i) can be done with SLAM by comparing to the offline map $M_{lidar}$. We omit the details and let the robot's current position be $P_{cur}$, and current angle be $A_{cur}$.

Task (ii) is done with the assistance of SAM (Segmentation Anything Model) \cite{kirillov2023segment}. The robot will rotate 360$^\circ$ at its current position $P_{cur}$ and take $\omega$ snapshots by its RGB-D camera. (In our implementation, $\omega=8$ and snapshots are taken every 45 degrees.) Let the RGB images be $I_{rgb}^1, I_{rgb}^2, \dots, I_{rgb}^\omega$ and depth images be $I_{dep}^1, I_{dep}^2, \dots, I_{dep}^\omega$. Each $I_{rgb}^i, i=1..\omega,$ is processed by an object detection model (such as YOLO \cite{cheng2024yolo}) to obtain a list of object bounding boxes. A unique ID is then assigned to each object. As the robot knows its orientation, duplicate objects appearing in two images can be easily recognized and removed. This potentially results in an object list $O_{list}$ such that each object is identifiable. Then, the bounding box image of each object $o_j \in O_{list}$ is cropped and sent to SAM to find its mask, named $s_j$.

Now, with $O_{list}$ and each $o_j$'s mask $s_j$, the next task is to map $o_j$ to the LiDAR map $M_{lidar}$. For each pixel $(x_p, y_p)$ within mask $s_j$, we compute:
\begin{equation}
    \begin{split}
        \Theta &= \frac{FoV_x}{w} \times (x_p - x_c) \\
        \Phi &= \frac{FoV_y}{h} \times (y_p - y_c)
        \\
        D_h &= d_p \times \cos{\Phi} \\
        \Delta x &= D_h \times \cos{(\Theta+A_{cur})} \\
        \Delta y &= D_h \times \sin{(\Theta+A_{cur})}
    \end{split}
    \label{eq:mapping}
\end{equation}
$\Theta$ and $\Phi$ are respectively the azimuth angle and elevation angle with respect to the robot's camera. 
$FoV_x$ represents the horizontal field of view, $FoV_y$ represents the vertical field of view, $w$ is the width of the sensor, $h$ is the height of the sensor, $(x_c, y_c)$ is the center of the image, and $d_p$ is the depth of pixel $(x_p, y_p)$ existing in a certain $(I_{rgb}^i, I_{dep}^i)$ pair. Then we compute $P_{cur} + (\Delta x, \Delta y)$ as the position of $(x_p, y_p)$ in $M_{lidar}$. 

By repeating the above process for all $(x_p, y_p) \in s_j$, we can map object $o_j$ onto $M_{lidar}$. By mapping each $o_j \in O_{list}$ onto $M_{lidar}$, we can obtain the online LiDAR map $\hat M_{lidar}$. 
An example is shown in Fig~\ref{fig:level1_example}.

\begin{figure} 
    \centering
    \includegraphics[width=0.35\textwidth]{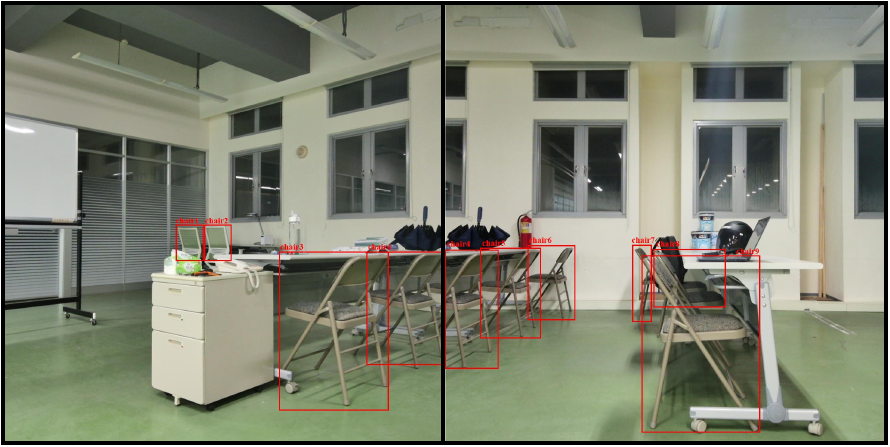}
    \caption{The annotated images $\hat I_{rgb}^x, x = 1..\omega,$ with object bounding boxes and object IDs shown in red.}
    \label{fig:bb}
\end{figure}

\subsection{Level-2 Mapping: Language-to-Object}

To initiate a mission, the user needs to exchange a sequence of dialogues, denoted by $D_1, D_2, \dots, D_k$, $k \ge 1$, with our conversation system. The responses of our conversation system to these dialogues are denoted by $R_1, R_2, \dots, R_k$, respectively. As our focus is positional ambiguity, we assume that $D_1$ contains time, position, target object, and action information of the mission, while $D_i, i \ge 2$, contain further information, if necessary, to clarify positional ambiguity. 

Fig. \ref{fig:dialogue} shows the architecture of the level-2 mapping. The first dialogue $D_1$ is sent to the LLM to interpret the user's intention. From the response $R_1$ of the LLM, we try to parse the time, position, target object, and required action information of the mission. If there is no ambiguity regarding the position, a mission can be created. Otherwise, the subsequent dialogues with the VLM will be exchanged. 

Suppose that the current dialogue with the user is $D_i, 2 \le i\le k$. We prompt the VLM with two inputs: (i) the text $D_i$ and (ii) the annotated images $\hat I_{rgb}^1, \hat I_{rgb}^2, \dots, \hat I_{rgb}^\omega$ obtained from the original RGB images $I_{rgb}^1, I_{rgb}^2, \dots, I_{rgb}^\omega$, respectively, by drawing bounding boxes and object ID tags on the images. Specifically, each $\hat I_{rgb}^x, x = 1.. \omega$, is obtained as follows. Recall that we already obtained the object list $O_{list}$ in which each object is unique. For each object $o_j$ appearing within $I_{rgb}^x$, we draw the bounding box of $o_j$ according to its coordinates. We also add a tag of $o_j$'s ID on top of the bounding box in the standard OpenCV style. This results in the annotated image $\hat I_{rgb}^x$. Several examples are shown in Fig.~\ref{fig:bb}. We will verify modern VLMs' capability in recognizing these annotated images in Sec. \ref{sec:exper}.

\begin{figure*}[t]
    \centering
    \begin{subfigure}{0.875\textwidth}
        \includegraphics[width=1\textwidth]{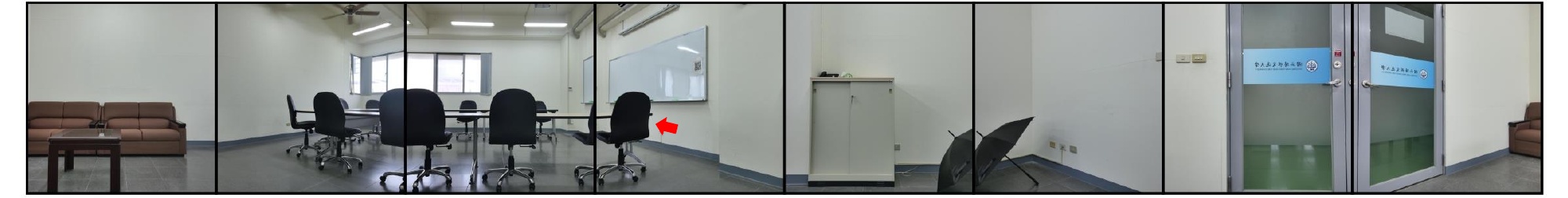}
        \caption{}
        \label{fig:ex-a}
    \end{subfigure}
    \begin{subfigure}{0.85\textwidth}
        \includegraphics[width=1\textwidth]{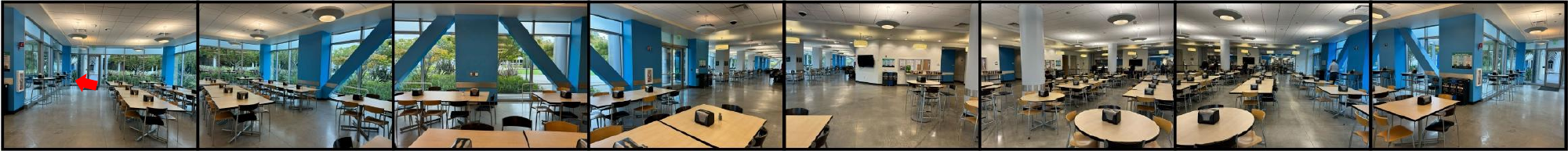}
        \caption{}
        \label{fig:ex-b}
    \end{subfigure}
    \caption{Examples of snapshots taken in (a) meeting room I and (b) cafeteria.}
\end{figure*}

Then the VLM is instructed to identify the precise bounding box and its unique ID that is intended by the user's dialogues. The VLM is instructed to respond in the format:
\begin{quote}
\texttt{The <object> is labeled as <object's ID> in the <which> image.}
\end{quote}
Then we parse the response $R_i$ of the VLM. If there is no positional ambiguity, a mission is created. Otherwise, the above process is repeated for the next dialogue $D_{i+1}$, until the maximal $k$ exceeds, in which case a failure is reported.

\subsection{Execution Module}

The above discussion addresses the initiation of one mission. According to the nature of the mission (immediate or scheduled), it will be placed in the mission scheduler. Then we use ROS commands to deploy the robot to execute these missions according to their schedules.

\section{Experiment Results}
\label{sec:exper}

\subsection{Vision-Dialogue Dataset}

The inputs to the VLM consist of $\omega = 8$ snapshots taken by the robot at a fixed position and a sequence of dialogues. To evaluate different VLMs' capability, we design a {\em Vision-Dialogue (VisDia)} dataset. Although the snapshots are easy to obtain, the dilemma is that the next dialogue $D_{i+1}$ should depend on the current reply $R_i$, making $D_{i+1}$ uncertain. To derive a fixed dataset, we designed a {\em gradual narrowing-down approach} when designing \emph{VisDia}. There are two types of dialogues:
\begin{itemize}
\item 
Type A:
The first dialogue $D_1$ refers to a specific object, but due to the robot's perspective or environmental complexity, the VLM may not be able to accurately identify the object. Therefore, subsequent dialogues $D_2, D_3, ..., D_k$ will provide additional information, progressively clarifying the object's identity until the robot can accurately determine the correct object.
\item 
Type B: Suppose that there is a set of $n_0$ objects of the same type in an environment. The first dialogue $D_1$ itself is ambiguous and covers a subset of $n_1$ objects ($n_1 < n_0$). Each subsequent dialogue $D_2, D_3, ..., D_k$ introduces additional constraints or details, progressively narrowing down the range of candidate objects, until the final dialogue $D_k$ precisely specifies a single target object ($n_k = 1$).
\end{itemize}
With the design, we are able to use fixed dialogue patterns to evaluate a VLM's capability in resolving positional ambiguity. Note that the length $k$ may vary in each data item.

The dataset was collected in 5 different spaces:
\begin{itemize}
\item 
Meeting room I: This space is about 50 $m^2$. The space is relatively simple and small, with 11 chairs, 1 table, 1 sofa set, 1 coffee table, 1 umbrella, 1 window, 1 cabinet, and 1 whiteboard. We designed 1 snapshot point and 10 type-A dialogue sequences.
\item 
Office: This space is about 150 $m^2$. The space is relatively large but simple, with 12 chairs, 4 tables, 5 windows, 1 cabinet, 4 whiteboards, and other office objects. We designed 1 snapshot point and 15 type-A dialogue sequences.
\item 
Meeting room II: This space is about 20 $m^2$. The space is relatively simple, with 10 chairs, 3 tables, and other everyday objects. We designed 3 snapshot points and 15 type-B dialogue sequences.
\item 
Classroom: This space is about 90 $m^2$. The space is somewhat complex, with approximately 40 chairs, 40 tables, classroom objects, and 2 sets of large whiteboards. We designed 1 snapshot point and 5 type-B dialogue sequences.
\item 
Cafeteria: This space is about 350 $m^2$. The space is highly complex, with approximately 150 chairs and 35 tables shown. We designed 4 snapshot points and 20 type-B dialogue sequences.
\end{itemize}


For example, the following type-A dialogue sequence is designed for the snapshots in Fig.~\ref{fig:ex-a} (the answer is the chair marked by a red arrow in the fourth snapshot).
\begin{quote}
\texttt{D1: Help me find the chair closest to the whiteboard.}
\\
\texttt{D2: It also needs to be the closest to the cabinet.}
\\
\texttt{D3: That chair is also next to the protruding wall.}
\\
\texttt{D4: It is also at the edge of the U-shaped table.}
\\
\texttt{D5: Finally, the chair should be directly in front of the white paper with the QR code.}
\end{quote}
The following type-B sequence is designed for Fig.~\ref{fig:ex-b} (the answer is the chair marked by a red arrow in the first snapshot).
\begin{quote}
\texttt{D1: Please go to the chair.}
\\
\texttt{D2: Hmm, I mean a high chair.}
\\
\texttt{D3: I think the high chair I need is left of the door and closest to it.}
\end{quote}

\subsection{Disambiguation Metrics}

To evaluate a VLM's disambiguation capability, we define two metrics for each dialogue type. Let $\mathcal{S}$ be the dataset and $s \in \mathcal{S}$ be a data item of type A with a dialogue length of $k$. We define {\em success rate} ($SR$) and {\em accuracy score} ($AS$) with respect to $s$ as follows:
\begin{equation}
    \begin{split}
        SR(s) &= (k-(\alpha -1)) / k \\
        AS(s) &=
        \begin{cases}
            \frac{1}{\alpha}
            \Sigma_{i=1}^\alpha 
            \frac{\textbf{1}(\text{found})}{\beta_i}, 
            & \text{if } \alpha \le k \\
            0                            , & \text{otherwise}
        \end{cases}
    \end{split}
\end{equation}

\begin{table}[t]
    \centering
    \caption{Quantitative evaluation of the disambiguation capability of GPT-4o. ($\lambda_{sr} = 0.8$ and $\lambda_{as} = 0.2$ for type-A dialogues; $\lambda_{ar} = 0.6$ and $\lambda_{ns} = 0.4$ for type-B dialogues.)}
    \begin{tabular}{c|cccccc}
                            &                    & $SR$ / $AR$ & $AS$ / $NS$ &  $T_{A/B}$ \\ \hline
    \multirow{2}{*}{Type-A} &   Meeting room I   & 0.636   &  0.79   &   0.667 \\
                            &   Office           & 0.866   &  0.835  &   0.86 \\ \hline
    \multirow{8}{*}{Type-B} &   Meeting room II  &   1     &  0.783  &   0.913 \\
                            &   Classroom-1      & 0.8     &  0.759  &   0.784 \\
                            &   Classroom-2      & 0.4     &  0.651  &   0.5 \\
                            &   Classroom-3      & 0.6     &  0.663  &   0.625 \\
                            &   Classroom-4      &   1     &  0.948  &   0.979 \\
                            &   Cafeteria-1      & 0.6     &  0.679  &   0.632 \\
                            &   Cafeteria-2      &   1     &  1      &   1.000 \\
                            &   Cafeteria-3      &   1     &  0.972  &   0.993 \\
    \end{tabular}
    \label{tab:FS}
\end{table}

Here, $\alpha$ is the number of dialogues consumed during the test. If $\alpha \le k$, the unique object is identified; otherwise, we assume that $\alpha = k+1$. So a smaller $\alpha$ results in a higher $SR(s)$. $\beta_i$ means the number of objects that the robot infers in response $R_i$. If the unique object is within the predicted set, \textbf{1}(\text{found}) is 1; otherwise, it is 0. So a smaller $\beta_i$ gives a higher $AS(s)$. We aggregate them into a total type-A score:

\begin{equation}
    T_A = \sum_{s \in \mathcal{S}}
    (\lambda_{sr} \times SR(s) + \lambda_{as} \times AS(s)) / |\mathcal{S}|
\end{equation}
where $\lambda_{sr}$ and $\lambda_{as}$ are weighting factors.

If $s \in \mathcal{S}$ is a type-B data item, we define the \emph{accuracy rate} ($AR$) and \emph{narrowing score} ($NS$) as follows:
\begin{equation}
    \begin{split}
        AR(s) &= 
        \begin{cases}
            1 , & \text{if } found \\
            0 , & \text{otherwise}
        \end{cases}
        \\
        NS(s) &= \frac{1}{\alpha}\sum_{i=1}^\alpha \frac{x_i \cap x'_i}{x_i \cup x'_i}
    \end{split}
\end{equation}
where $x_i$ is the set of actual objects in dialogue $D_i$ and $x'_i$ is the set of predicted objects by the model in $R_i$. We aggregate them into a total type-B score:
\begin{equation}
    T_B = \sum_{s \in \mathcal{S}}
    (\lambda_{ar} \times AR(s) + \lambda_{ns} \times NS(s)) / |\mathcal{S}|
\end{equation}
where $\lambda_{ar}$ and $\lambda_{ns}$ are weighting factors.

Based on the above metrics, we show our quantitative evaluation results in Table~\ref{tab:FS}. These experiments were conducted by using GPT-4o as the LLM and VLM. For type-A dialogues, we observe that in the office setup, the $SR$, $NS$, and $T_A$ scores are all slightly higher than those of the meeting room I setup. We suspect the reason to be that the types and numbers of objects in these two spaces are similar. However, the office space is relatively larger, making this task relatively easier for the VLM to solve. For type-B dialogues, the meeting room II has consistently high scores since its setup is relatively simple. In contrast, we observe inconsistent scores under different cases for the classroom and cafeteria setups, which are more complex. For example, the scores of classroom-4, cafeteria-2, and cafeteria-3 are pretty high, while the scores for the other cases are much lower. This is due to the complexities of these spaces, including numbers, types, sizes, and diversities of the objects in these spaces. It is to be noted that due to our formulations, $T_A$ scores will generally be lower than $T_B$ scores.

\begin{table}[t]
    \centering
    \caption{Error analyses of level-1 mapping.}
    \begin{tabular}{c|cc}
                              & Meeting room I & Office \\ \hline
    Mean Error $(m)$ &          0.657 & 0.983  \\
    Standard Deviation $(m)$ &  0.566 & 1.159  \\
    Min Error $(m)$ &           0.058 & 0.121  \\
    Max Error $(m)$ &           1.848 & 4.923  \\
    \end{tabular}
    \label{tab:distance_error}
\end{table}

\begin{figure}[t]
    \centering
    \includegraphics[width=0.4\textwidth]{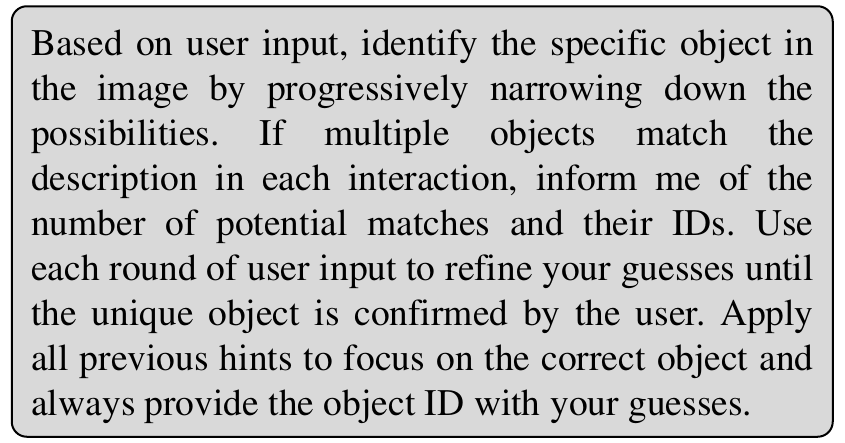}
    \caption{The instruction to set up GPT-4o.}
    \label{fig:instruction}
\end{figure}

\subsection{Object Mapping Accuracy}

Here, we test level-1 mapping accuracy for the two environments with type-A dialogues (i.e., meeting room I and office). We omit the cases of type-B dialogues as level-1 mapping is irrelevant to the type of dialogues. Using Eq. \ref{eq:mapping}, we use the average depth of an object to estimate its position in the LiDAR map $M_{lidar}$. We then calculate the error between the predicted position to the ground truth center of the object. The results reflect the accuracy of SAM and hardware limitations. As Table \ref{tab:distance_error} shows, the errors are within an acceptable range. Since the office space is much larger, it also has a larger error than meeting room I.

\subsection{Implementation Details and Demos}

Our robot testbed was developed based on ROS Melodic. The main control board was NVIDIA Jetson Nano, and the operating system used was Ubuntu 18.04 LTS. The robot was equipped with a LiDAR and two RGB-D cameras. The LiDAR was for constructing $M_{lidar}$ and positioning the robot. The first RGB-D camera, ORBBEC Astra S, was used for obstacle avoidance and was installed at the bottom of the robot. The second RGB-D camera, ORBBEC Femto Bolt, was installed at the top of the robot for sensing the environment. We used YOLOv8~\cite{varghese2024yolov8} for object detection and SAM~\cite{kirillov2023segment} for object segmentation. We adopted GPT-4o~\cite{achiam2023gpt} as both the LLM and the VLM in our level-2 mapping module. The robot interacted with GPT-4o through its APIs. We have also tried other models; refer to Sec. \ref{sec:conclusion} for more details.

Note that before prompting GPT-4o with dialogues $D_1, D_2, \dots, D_k$, we need to give an instruction to set up the mission and context for GPT-4o to respond. The instruction is shown in Fig.~\ref{fig:instruction}

\begin{figure}[t]
    \centering
    \includegraphics[width=0.47\textwidth]{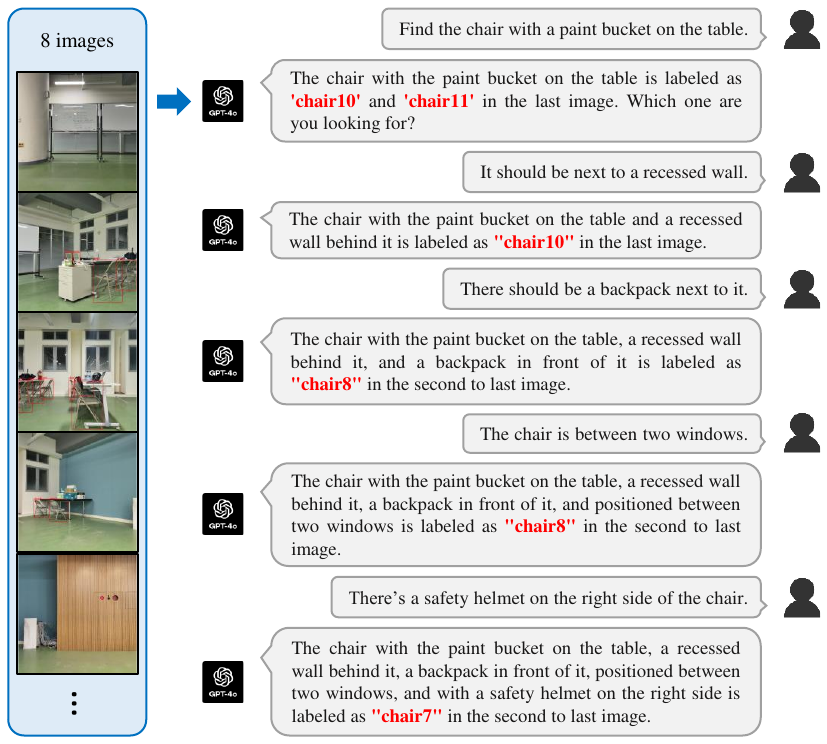}
    \caption{A successful prompting example (type-A). The target object, chair7, is with a paint bucket on the table, near a recessed wall, with a backpack beside it. It sits between two windows, and a safety helmet is on its right. Type-A is to test how much information is needed for the model to correctly identify the object, as there is a unique answer starting from the beginning of the conversations.}
    \label{fig:ex-success}
\end{figure}

In Fig.~\ref{fig:ex-success}, we demonstrate a successful prompting example, where the dialogues are of type A. A failure example is demonstrated in Fig.~\ref{fig:ex-failure}, where the dialogues are of type B.

\section{Conclusions and Limitations}
\label{sec:conclusion}

In this work, we have addressed an interesting positional ambiguity issue that is rooted at the imprecision of most natural language commands. Although modern computer vision techniques can recognize and position objects with high accuracy, a robot may not follow human commands and navigate to the right position if such ambiguities are not well clarified. This problem may become more serious in indoor environments where many objects may be of the same types. The work contributes in proposing the use of multiple dialogues to clarify amibguities, developing a 2-level mapping framework, leveraging modern LLM and VLM for text and vision mapping, defining type-A and type-B dialogues in such scenarios, deriving the VisDia dataset, and validating the above issues via a ROS-based robot. In particular, our 2-level mapping separates language-to-object alignment and object-to-LiDAR map alignment, which could be generalized to resolving other tasks in the future.

Below, we point out some limitations in our current work. Our current testbed adopted GPT-4o to serve as the LLM and VLM. Our tests revealed that GPT-4o is the only VLM capable of handling multiple image inputs at a time. Most other VLMs accept and process only one image per conversation, which do not meet our needs. Providing multiple images as visual inputs is necessary unless a fisheye camera is utilized. To address this, we discuss two alternatives:

\begin{figure}[t]
    \centering
    \includegraphics[width=0.47\textwidth]{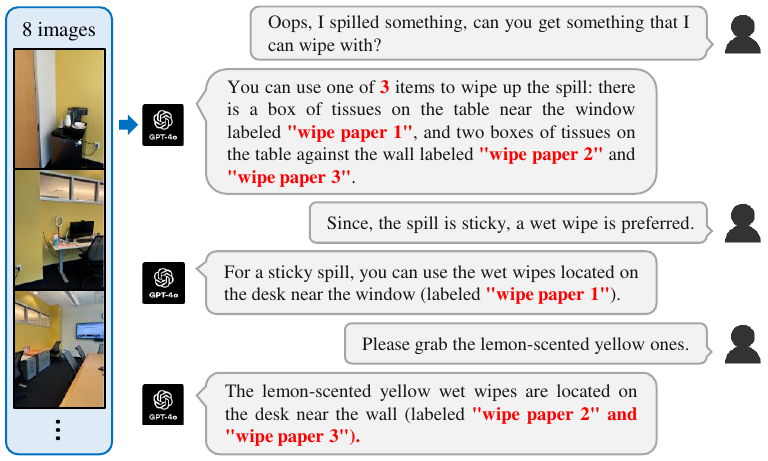}
    \caption{A failure prompting example (type-B). The target object that we are looking for is `wipe paper 3'. Type-B is designed to test whether the model can interpret ambiguous human commands and among those objects covered by an ambiguous command how many are correctly identified.}
    \label{fig:ex-failure}
\end{figure}

\begin{itemize}
\item 
Concatenating multiple images into a single large image:
This can be in a form of a long strip or a grid, and the concatenated image serves as the input to the VLM. For example, LLaVA1.5~\cite{liu2024improved} restricts its image size to $336\times 336$ pixels. If the input image exceeds this size, it will be automatically resized. This results in images that are too blurry and consequently affects its recognition capability.
\item 
Supplying a sequence of images in multiple rounds of conversations: For instance, we can send in 8 images sequentially. This can lead to an issue where multiple rounds of interactions occur before the user's actual dialogues, causing the catastrophic forgetting problem and significantly impacting the quality of subsequent responses from the VLM.
\end{itemize}
Due to the above limitations, our current experiments were primarily conducted on GPT-4o. Using smaller, local VLMs is still prohibitive at this moment.

\bibliographystyle{IEEEtran}
\bibliography{IEEEfull}

\end{document}